# Bayesian Physics-Informed Neural Network for the Forward and Inverse Simulation of Engineered Nano-particles Mobility in a Contaminated Aquifer

Shikhar Nilabh, Fidel Grandia

Amphos 21 Consulting S.L., c/Veneçuela 103, 08019 Barcelona, Spain

Globally, there are many polluted groundwater sites that need an active remediation plan for the restoration of local ecosystem and environment. Engineered nanoparticles (ENPs) have proven to be an effective reactive agent for the in-situ degradation of pollutants in groundwater. While the performance of these ENPs has been highly promising on the laboratory scale, their application in real field case conditions is still limited. The complex transport and retention mechanisms of ENPs hinder the development of an efficient remediation strategy. Therefore, a predictive tool to comprehend the transport and retention behavior of ENPs is highly required. The existing tools in the literature are dominated with numerical simulators, which have limited flexibility and accuracy in the presence of sparse datasets and the aquifer heterogeneity. This work uses a Bayesian Physics-Informed Neural Network (B-PINN) framework to model the nano-particles mobility within an aquifer. The result from the forward model demonstrates the effective capability of B-PINN in accurately predicting the ENPs mobility and quantifying the uncertainty. The inverse model output is then used to predict the governing parameters for the ENPs mobility in a small-scale aquifer. The research demonstrates the capability of the tool to provide predictive insights for developing an efficient groundwater remediation strategy.

## Introduction

With more than a billion people lacking access to clean drinking water, the water shortage is a rapidly increasing concern [1]. The widespread presence of contamination in aquifers poses a great threat to the human health, environmental quality and socioeconomic development [2], [3]. Furthermore, the groundwater quality is intricately linked to the climate change. The persistence of groundwater pollution can lead to the degradation of soil quality [4], forestry [5] and even coastal ecosystem [6]; thus accelerating the climate change processes. Additionally, groundwater pollution can release toxic gases, such as methane, into the atmosphere, contributing to the overall increase in greenhouse gas emissions. Implementing groundwater remediation techniques in a contaminated aquifer is essential for the well-being of the local and global ecosystems [7].

Several remediation techniques have been developed in the literature for the restoration of groundwater resources [8]–[10]. In the recent years, injection of ENPs in a contaminated aquifer has proven to be highly efficient in the groundwater remediation [11], [12]. While extensive research has been done in developing and testing these ENPs at laboratory scale, their field scale injection for groundwater remediation is still limited [13]. The limitation of ENPs application is attributed to its highly complex transport and retention behavior governed by heterogeneities in the porous media [14]. While several numerical and data-driven tools have been proposed in the literature for predicting the behavior of ENPs, their application is often challenged by data sparsity and heterogeneity in the aquifer sand [15]. A Physics-Informed Neural Network (PINN) can overcome these challenges by embedding the knowledge of governing equations into the learning method of a neural network [16]. Additionally, the noise in the dataset due to heterogeneities in the porous media can be characterized by implementing a Bayesian inference. However, the application of point estimate or Bayesian PINNs in the subsurface flow and transport is still rare and have been only limited to point groundwater flow and contaminant transport [17], [18]. To our best of knowledge, this paper is the first to implement Bayesian Physics-Informed Neural Network (BPINN) for the simulation of ENPs mobility´s parameters and their uncertainty quantification. This approach intends to bridge the gap between the testing of ENPs in laboratories and their field application. In this regard, a two-fold objective of this work is defined: a) development and verification of forward B-PINN model simulating nano-particle's transport and retention in a 1-Dimensional column-filled sand. b) inverse modeling using B-PINN for estimating the physiochemical properties of ENPs. The purpose of this research is to show that the tool can be used to understand the mobility of ENPs; using it in a real-world case study is beyond the scope of this work.

## Methodology

The nano-particle´s fate in the aquifer is governed by transport due to groundwater flow and retention due to particle-sand interaction. The equation 1 and 2 represent the modified Advection Dispersion equation for the ENPs transport and retention in a 1D column-filled saturated sand representing a small scale aquifer.

$$\frac{\partial(\theta c)}{\partial t} + \frac{\partial(s)}{\partial t} = -\nabla(vc) + \nabla((D_e + \propto_L v)\nabla c) \tag{Eq. 1}$$

$$\frac{\partial(s)}{\partial t} = \theta k_a c - k_d s \tag{Eq. 2}$$

$$c(0,t) = \left(\frac{1}{1+e^{-0.02(t-500)}}\right) \times \left(\frac{1}{1+e^{0.02(t+4100)}}\right) \tag{Eq. 3}$$

$$(D_e + \propto_L v)\nabla c(1,t) = 0 \tag{Eq. 4}$$

Where $\theta$ is the porosity, $D_e$ is molecular dispersivity (m²·s⁻¹), $\propto_L$ is dispersivity (m), $k_a$ (s⁻¹) and $k_d$ s⁻¹ is the attachment and the detachment coefficient of ENPs respectively, c (kg·m⁻³) and s (kg·m⁻³) are aqueous and retained concentration of ENPs respectively. The equation 3 and 4 represents the Dirichlet and Neumann boundary condition assigned at the each end of the sand column. The initial concentration of ENPs, $c(x, 0)$, is considered to be zero.

Two fully connected neural network are considered to approximate the aqueous and retained concentration of ENPs (Figure 1) Each of the network consists of 6 hidden layers each with 50 neurons. 3000 collocation points are used for enforcing boundary condition where as, with Lattice Hypercube Sampling, 15000 collocation points are selected for enforcing equation 1 and 2. Sigmoid activation function is used for incorporating non-linearity in the neural network. The weights used in the loss function has been taken from the previous studies in the literature [19].

In the subsequent step, the inverse model is developed to estimate the physiochemical properties of ENPs. For this a synthetic dataset of breakthrough curve and retention profile is used. The data is obtained in Comsol by considering the physiochemical parameters $k_a$ and $k_d$ to be 0.0008 s⁻¹ and 0.0001 s⁻¹. A Guassian noise $\mathcal{N}(0, 0.001^2)$ is added in the synthetic data to mimic the noise likely to be found during the experimental set-up. All the other parameters have been kept the same as the forward model set-up.

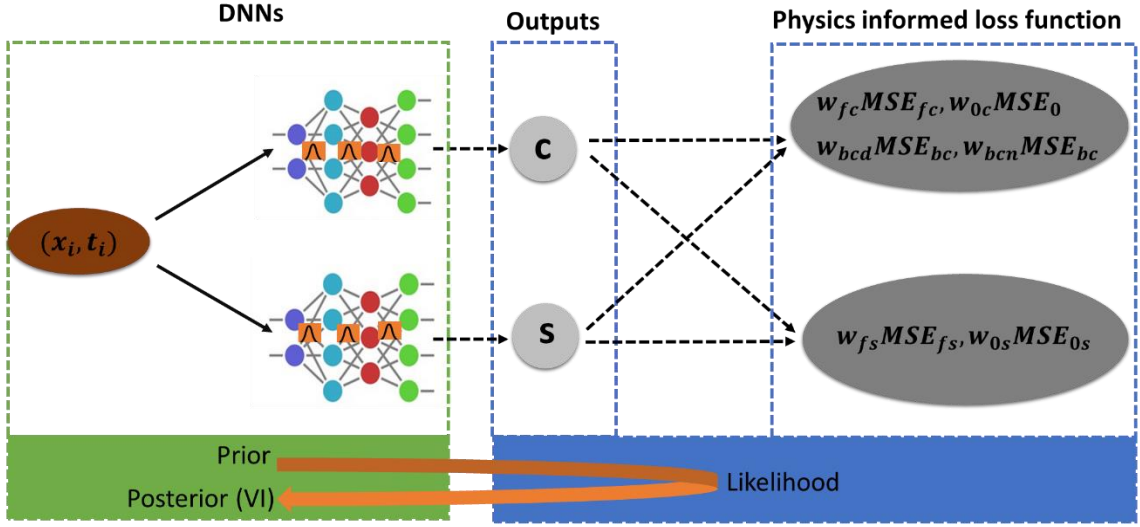

*Figure 1: Schematics of the B-PINN framework. Two Deep Neural Networks take time and spatial points as input to predict concentration of aqueous and retained ENPs. The loss function J consists of 6 Mean square error term (MSEs) corresponding to the boundary and PDE equations. The posterior distribution of network parameters are estimated using their prior distribution and likelihood of the observations.*

The vector of the network parameter $\boldsymbol{\theta}$ is considered to have a gaussian prior distribution $P(\boldsymbol{\theta})$ with mean 0 and standard deviation of 0.001. The likehlihood can be given by Equation 5.

$$P(\mathcal{D} \mid \boldsymbol{\theta}) = P(\mathcal{D}_u \mid \boldsymbol{\theta})P(\mathcal{D}_f \mid \boldsymbol{\theta})P(\mathcal{D}_b \mid \boldsymbol{\theta}) \tag{Eq. 5}$$

Where $\mathcal{D}$ is the available data in the form of observation $\mathcal{D}_u$, boundary conditions $\mathcal{D}_b$ and $\mathcal{D}_f$ as residuals. The available data are considered to have intrinsic noise which has gaussian distribution $\epsilon_u \sim \mathcal{N}(0, 0.01^2), \epsilon_f \sim \mathcal{N}(0, 0.00000001^2)$ and $\epsilon_b \sim \mathcal{N}(0, 0.001^2)$. Relatively small noise is considered for residual data $\mathcal{D}_f$ due to the slow dynamics of ENP´s advection.

Variation inference is used to find a parametric approximation to the true posterior distribution. It is achieved by introducing a Guassian distribution over the model's parameters and minimizing the difference, measured by the Kullback-Leibler (KL) divergence, between the approximating distribution and the true posterior. For this, 20000 iterations of Adam optimization are used to estimate the mean and standard deviation of network parameters.

## Result

### Model verification

In the first stage, a forward model is generated with B-PINN and compared with the results of Comsol model. 50 network parameters were sampled from the approximated distribution of the posterior. Each sampled value of the parameters results in the unique prediction of the aqueous and retained concentration of ENPs by the neural network shown by colored curves (Figure 2a, Figure 2b). The benchmark data obtained from Comsol falls (represented by black curve) reasonably within the range of the neural network´s output (represented by the colored curves in Figure 2). The breakthrough curve in Figure 2 (a) shows that the result of B-PINN model is relatively smoother in the region where Comsol model has undergone overshooting and undershooting. Overall, result demonstrates the effective capability of B-PINN in predicting the mobility and retention of ENPs in saturated sand.

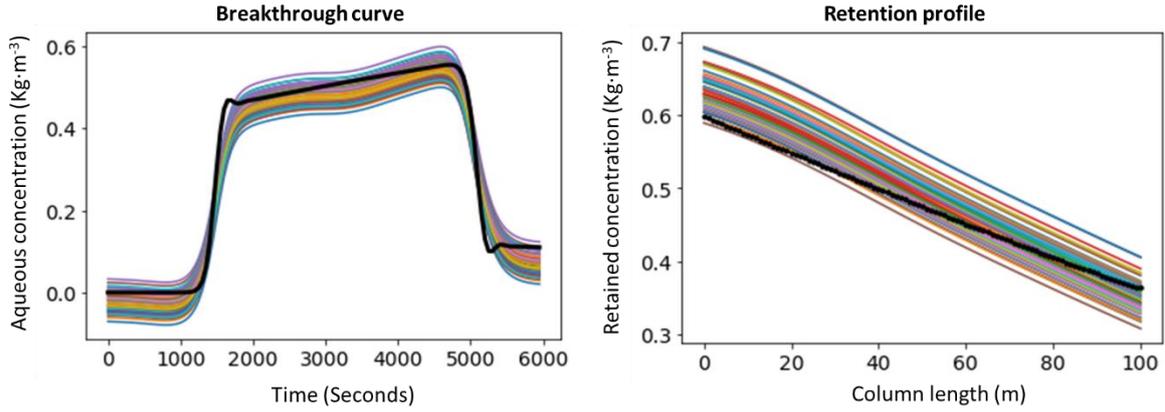

*Figure 2: Result comparison of B-PINN and Comsol model; (a) breakthrough curve (b) retention. The black curve represents the benchmark data from Comsol whereas the range of colored curves represent the model´s output based on each of the sampled network´s parameters.*

Inverse modeling

50 network parameters were sampled from the approximated distribution of the posterior to obtain the output of the neural network. Figure 3 (a) and 3 (b) shows the breakthrough curve and retention profile of the ENPs simulated using B-PINN (represented by the colored curves) and its close fit with the training dataset generated using Comsol (represented by black curve). The result highlights model's ability to simulate the mobility dynamics of ENPs with reasonable accuracy. The implementation of Bayesian inference techniques such as Hamiltonian Monte Carlo and Monte Carlo Dropout instead of Variational Inference may further increase the model´s accuracy.

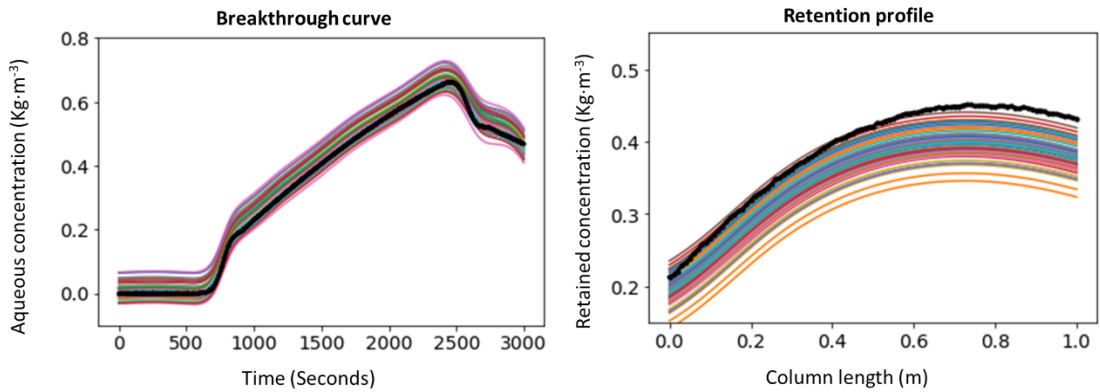

*Figure 3: Result comparison of inverse P-PINN model and the training dataset generated using Comsol; (a) breakthrough curve for ENPs (b) Retention profile for ENPs. The black curve represents the synthetic data from Comsol whereas the range of colored curves represent the model´s output based on each of the sampled network´s parameters.*

The model estimates the gaussian distribution of the ENPs properties, $k_a$ and $k_d$ to be $\mathcal{N}(0.00076, 0.00000614^2)$ and $\mathcal{N}(0.00012, 0.00000612^2)$ respectively. The mean of $k_a$ and $k_d$ estimated by the model is very close to the values used for the generation of the synthetic data. The uncertainties represented by the standard deviation have a relatively lower magnitude compared to the mean. However, the uncertainty can possibly be higher in the real experimental case. The model result demonstrates its ability to not only estimate the physiochemical properties of the ENPs but also the uncertainties associated in the estimation process.

# Conclusion

This research aims to develop a modeling tool that can provide useful predictions for groundwater remediation planning. The study found that a forward and inverse B-PINN tool is effective in analyzing the transport and retention of nanoparticles in a small aquifer. The forward model has been verified using Comsol Multiphysics and has a minimal RMSE of $1.6\times10^{-5}$. The inverse model, based on the same verified framework, accurately determines the parameters for nanoparticle transport with reasonable accuracy. The uncertainty quantification with B-PINN can be instrumental in simulating the real case where the noise due to sand heterogeneity is significant. This tool can bridge the gap between laboratory testing and field implementation oof ENPs and can be used in a two-step process for developing an effective groundwater remediation plan. First, the inverse model can be used to estimate parameters based on experimental data. Then, a forward model can be created using these parameters to provide relevant predictions for groundwater remediation.